\documentclass{article}

\usepackage[final, nonatbib]{neurips_data_2023}

\usepackage[utf8]{inputenc} % allow utf-8 input
\usepackage[T1]{fontenc}    % use 8-bit T1 fonts
\usepackage[pagebackref=true,breaklinks=true,colorlinks,bookmarks=false]{hyperref}
\usepackage{url}            % simple URL typesetting
\usepackage{booktabs}       % professional-quality tables
\usepackage{amsfonts}       % blackboard math symbols
\usepackage{nicefrac}       % compact symbols for 1/2, etc.
\usepackage{microtype}      % microtypography
\usepackage{epsfig}
\usepackage{graphicx}
\usepackage{amsmath}
\usepackage{amssymb}
\usepackage{booktabs}
\usepackage{threeparttable}
\usepackage[table,xcdraw]{xcolor}
\usepackage{multirow}
\usepackage{subcaption}
\usepackage[capitalize]{cleveref}
\usepackage{wrapfig}
\usepackage{xspace}

\newcommand{\papername}{EgoTracks\xspace} 

\makeatletter
\DeclareRobustCommand\onedot{\futurelet\@let@token\@onedot}
\def\@onedot{\ifx\@let@token.\else.\null\fi\xspace}
\def\eg{\emph{e.g}\onedot}

\newenvironment{tight_itemize}{
\begin{itemize}%[leftmargin=*]
  \setlength{\topsep}{0pt}
  \setlength{\itemsep}{2pt}
  \setlength{\parskip}{0pt}
  \setlength{\parsep}{0pt}
}{\end{itemize}}

\newenvironment{tight_enumerate}{
\begin{enumerate}%[leftmargin=*]
  \setlength{\topsep}{0pt}
  \setlength{\itemsep}{2pt}
  \setlength{\parskip}{0pt}
  \setlength{\parsep}{0pt}
}{\end{enumerate}}

\begin{document}

\title{EgoTracks: A Long-term Egocentric Visual Object Tracking Dataset}

% The \author macro works with any number of authors. There are two commands
% used to separate the names and addresses of multiple authors: \And and \AND.
%
% Using \And between authors leaves it to LaTeX to determine where to break the
% lines. Using \AND forces a line break at that point. So, if LaTeX puts 3 of 4
% authors names on the first line, and the last on the second line, try using
% \AND instead of \And before the third author name.

% \author{%
%   Hao Tang \\
%   Meta AI\\
%   \texttt{haotang@meta.com} \\
%   % examples of more authors
%   \And
%   Kevin J Liang \\
%   Meta AI \\
%   % Address \\
%   \texttt{kevinjliang@meta.com} \\
%   \AND
%   Kristen Grauman \\
%   Meta AI and UT Austin \\
%   % Address \\
%   \texttt{grauman@meta.com} \\
%   \And
%   Matt Feiszli \\
%   Meta AI \\
%   % Address \\
%   \texttt{mdf@meta.com} \\
%   \And
%   Weiyao Wang \\
%   Meta AI \\
%   % Address \\
%   \texttt{weiyaowang@meta.com} \\
% }
\author{%
  Hao Tang$^1$, Kevin J Liang$^1$, Kristen Grauman$^{1,2}$, Matt Feiszli$^{1,*}$, Weiyao Wang$^{1,*}$ \\
  Meta AI$^1$, UT Austin$^2$, Equal Contribution$^*$\\
  \texttt{\{haotang, kevinjliang, grauman, mdf, weiyaowang\}@meta.com} \\
}

% \begin{document}

\maketitle

\begin{abstract}
\vspace{-1mm}
Visual object tracking is key to many egocentric vision problems.  However, the full spectrum of challenges of egocentric tracking faced by an embodied AI is underrepresented in many existing datasets, which tend to focus on short, third-person videos.
Egocentric video has several distinguishing characteristics from those commonly found in past datasets: 
% Egocentric vision often involves large camera motions, which causes objects to regularly exit and re-enter the frame, and hand interactions with objects lead to many occlusions, as well as widely different points of view, scale, or object states.
frequent large camera motions and hand interactions with objects commonly lead to occlusions or objects exiting the frame, and object appearance can change rapidly due to widely different points of view, scale, or object states.
Embodied tracking is also naturally long-term, and being able to consistently (re-)associate objects to their appearances and disappearances over as long as a lifetime is critical.
Previous datasets under-emphasize this re-detection problem, and their ``framed'' nature has led to adoption of various spatiotemporal priors that we find do not necessarily generalize to egocentric video. 
We thus introduce \papername, a new dataset for long-term egocentric visual object tracking. 
Sourced from the Ego4D dataset, \papername presents a significant challenge to recent state-of-the-art single-object trackers, which we find score more poorly on our new dataset than existing popular benchmarks, according to traditional tracking metrics.
We further show improvements that can be made to the STARK tracker to significantly increase its performance on egocentric data, resulting in a baseline model we call EgoSTARK.
We publicly release our annotations and benchmark (\url{https://github.com/EGO4D/episodic-memory/tree/main/EgoTracks}), hoping our dataset leads to further advancements in tracking.
\end{abstract}

% %%%%%%%%% BODY TEXT
% Fully leaning into egocentric aspect of tracking

\vspace{-2mm}
\section{Introduction}
\label{sec:intro}
\vspace{-1mm}

\begin{figure}[t]
\hfill
\includegraphics[width=0.9\columnwidth]{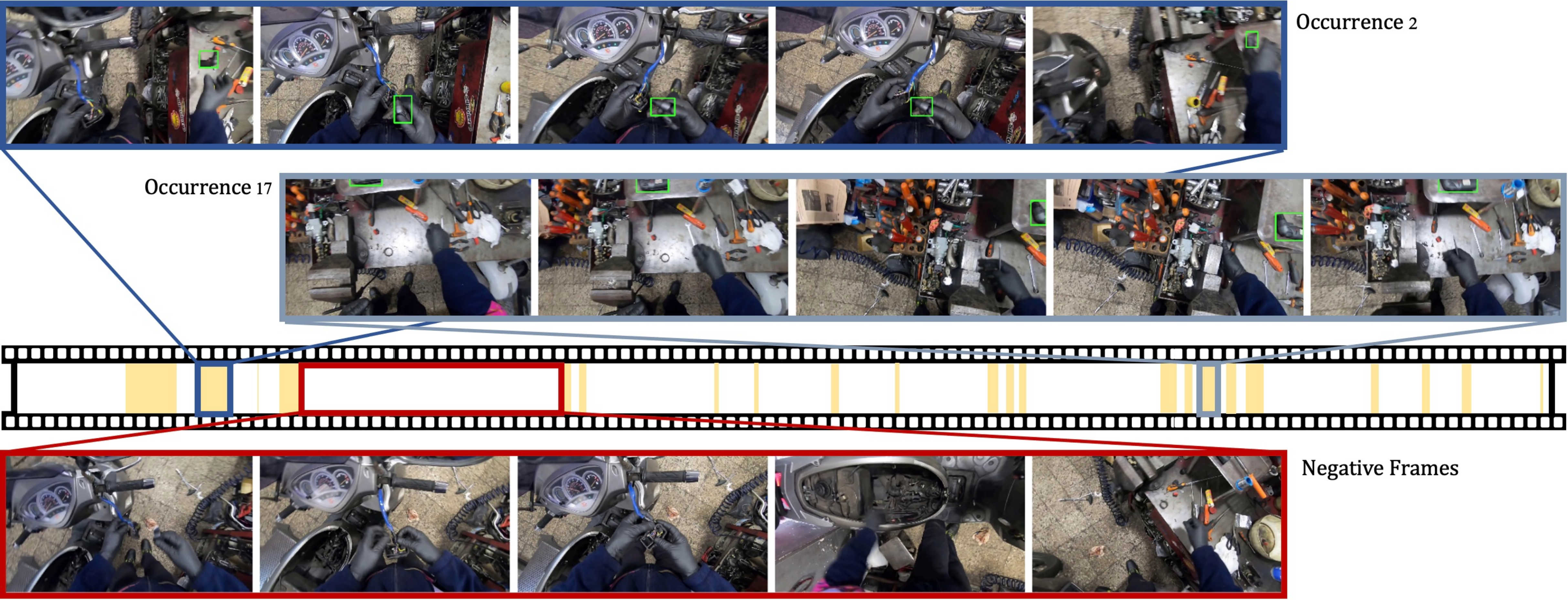}
% \caption{Word cloud of scenarios and activities.}
\hfill
\vspace{-1mm}
\caption{A video from the proposed \papername dataset, with yellow clip segments marking when the object (\texttt{blowtorch}) is visible. Note the frequent disappearances and reappearances of the object over an 8 minute video, with lengthy absences, necessitating re-detection to track accurately without false positives. The egocentric nature of the video includes the camera-wearer interacting with the object (Occurrence 2), resulting in significant hand occlusions and dramatic changes in pose.
}
\label{fig:teaser}
\vspace{-4mm}
\end{figure}

First-person or ``egocentric'' computer vision aims to capture the real-world perceptual problems faced by an embodied AI; it has
% \matt{Maybe introduce the idea of an embodied AI here quickly for definition and motivation?  Something like "First-person or egocentric video aims to capture the real-world perceptual problems faced by an embodied AI.  Egocentric vision has drawn strong interest..."  This also gives some leeway to refer back to the problem as fundamental compared to 3-p video.} 
drawn strong recent interest as an underserved but highly relevant domain of vision, with important applications ranging from robotics~\cite{savva2019habitat,duan2022survey} to augmented and mixed reality~\cite{azuma1997survey,speicher2019mixed,grauman2022ego4d}.
Visual object tracking (VOT), long a fundamental problem in vision, is a core component to many egocentric tasks, including tracking the progress of an action or activity, (re-)association of objects in one's surroundings, and predicting future states of the environment.
Yet, while the VOT field has made many significant advancements over the past decade, tracking in egocentric video remains underexplored.
This lack of attention is in large part due to the absence of a large-scale egocentric tracking dataset for training and evaluation.
While the community has proposed a number of popular tracking datasets in recent years, including OTB~\cite{wu2013online}, TrackingNet~\cite{muller2018trackingnet}, GOT-10k~\cite{huang2019got}, and LaSOT~\cite{fan2019lasot}, we find that the strong performance that state-of-the-art trackers achieve on these benchmarks does not translate well to egocentric video, thus establishing a strong need for such a tracking dataset.

We attribute this performance gap to the many unique aspects of egocentric views compared to the more traditional third-person views of previous datasets.
In contrast to intentionally ``framed'' video, egocentric videos are often uncurated, meaning they tend to capture many attention shifts between activities, objects, or locations.
Due to the first-person perspective, large head motions from the camera wearer often result in objects repeatedly leaving and re-entering the field of view; similarly, hand manipulations of objects~\cite{shan2020understanding} leads to frequent occlusions, rapid variations in scale and pose, and potential changes in state or appearance.
Furthermore, egocentric video tends to be long (sometimes representing the entire life of an agent or individual), meaning the volume of the aforementioned occlusions and transformations scales similarly.
These characteristics all make tracking objects in egocentric views dramatically more difficult than scenarios commonly considered in prior datasets, and their absence represents an evaluation blindspot.

Head motions, locomotion, hand occlusions, and temporal length lead to several challenges. 
First, frequent object disappearances and reappearances causes the problem of \textit{re-detection} within egocentric tracking to become especially critical.
Many previous tracking datasets primarily focus on short-term tracking in third-person videos, providing limited ability to evaluate many of the challenges of long-term egocentric tracking due to low quantity and length of target object disappearances.
As a result, competent re-detection is not required for strong performance, leading many recent short-term trackers to neglect it, instead predicting a bounding box for every frame, which may lead to rampant false positives or tracking the wrong object.
Additionally, the characteristics of short-term third-person video have also induced designs relying on gradual changes in motion and appearance.
As we later show (Section~\ref{sec:spacetimepriors}), many of the motion, context, and scale priors made by previous short-term tracking algorithms fail to transfer to egocentric video.

\begin{wrapfigure}{r}{.5\linewidth}
% \vspace{-4mm}
\centering
\includegraphics[width=0.45\columnwidth]{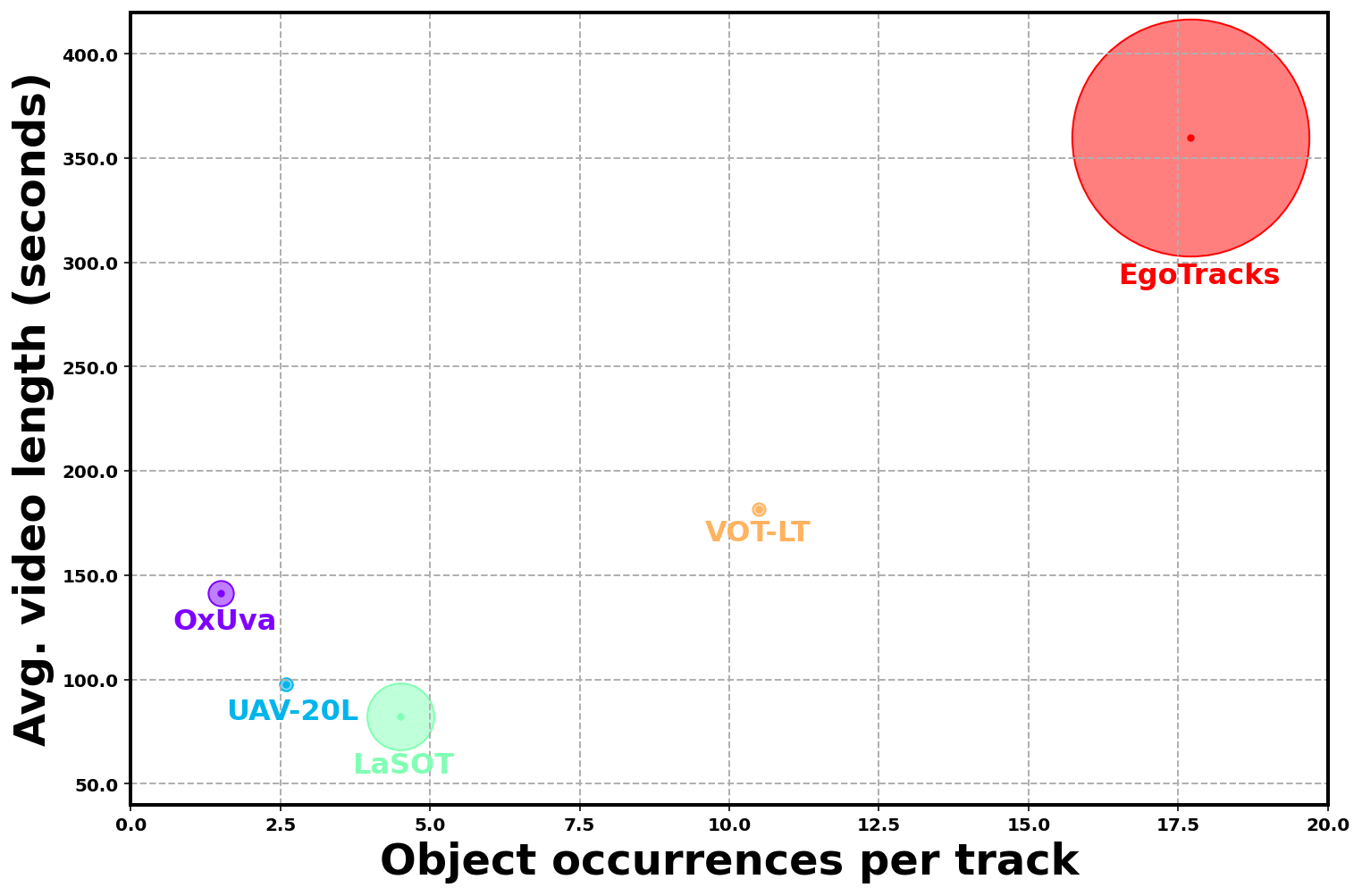}
% \caption{Word cloud of scenarios and activities.}
\caption{\papername is an order of magnitude larger than past long-term VOT datasets, with significantly more tracks (circle area) and object disappearances/appearances in longer videos.
% \matt{Looks good -- give another sentence explaining the chart in words.  Also "\# of occurrences" doesn't specify what's occurring -- maybe "object occurrences per track" or similar, and/or just be more explicit in the caption.} 
% Circle area indicates total number of tracks.
}
\label{fig:scale_comp}
% \vspace{-5mm}
\end{wrapfigure}
Notably, re-detection, occlusions, and longer-term tracking have long been recognized as difficult for VOT as a field, leading to recent benchmark construction efforts \cite{lukevzivc2018now,dai2020high,moudgil2018long,valmadre2018long,huang2020globaltrack,voigtlaender2020siam} emphasizing these aspects.
We argue that egocentric video provides a natural source for these challenges at scale while also representing a highly impactful application for tracking, therefore constituting a significant opportunity.
We thus present \textbf{\papername}: a large-scale long-term egocentric visual object tracking dataset for training and evaluating long-term trackers. 
Seeking a realistic challenge, we source videos from Ego4D~\cite{grauman2022ego4d}, a large-scale dataset consisting of unscripted, in-the-wild egocentric videos of daily-life activities.
The result is a large-scale dataset to evaluate the tracking and re-detection ability of SOT models, with more than 22,028 tracks from 5708 average 6-minute videos.
This constitutes the first large-scale dataset for visual object tracking in egocentric videos in diverse settings, providing a new, significant challenge compared with previous datasets.

We perform a thorough analysis of our new dataset and its new characteristics relative to prior benchmarks, 
demonstrating its difficulty and the need for further research to develop trackers 
capable of handling long-term egocentric vision. 
Our experiments reveal remaining open problems and insights towards promising directions in egocentric tracking. Leveraging these intuitions, we propose multiple simple yet effective changes, such as adjustment of spatiotemporal priors, egocentric data finetuning, and combining multiple templates. We apply these proposed strategies on a state-of-the-art (SOTA) STARK tracker~\cite{yan2021learning}, training a strong tracker dedicated to long-term egocentric tracking: \textbf{EgoSTARK}. We hope Ego-STARK can serve as a strong baseline and facilitate future research. 

% while also showing that a few improvements to the state-of-the-art STARK tracker~\cite{yan2021learning} can result in significant performance improvements on long-term tracking metrics.

We make the following contributions: 
\vspace{-2mm}
\begin{tight_enumerate}
    \item  We present \papername, the first large-scale long-term object tracking dataset with diverse egocentric scenarios. We analyze its uniqueness in terms of evaluating the re-detection performance of trackers. 
    \item We conduct comprehensive experiments to understand the performance of many state-of-the-art trackers on the \papername validation set and observe that due to the biases and evaluation blindspots of existing third-person datasets, they tend to struggle.
    \item We perform an analysis of what makes a good tracker for long-form egocentric video. Applying these learnings to the STARK tracker~\cite{yan2021learning}, we produce a strong baseline we call EgoSTARK, which achieves significant improvements (+15\% F-score) on \papername. 
\end{tight_enumerate}
\vspace{-2mm}

\vspace{-2mm}
\section{Related work}
\vspace{-2mm}
\subsection{Visual object tracking datasets}
\vspace{-2mm}
Visual object tracking studies the joint spatial-temporal localization of objects in videos. 
% There are two main categories: multiple object tracking (MOT) and single object tracking (SOT). 
From a video and a predefined taxonomy, multiple object tracking (MOT) models simultaneously detect, recognize, and track multiple objects. For example, MOT~\cite{milan2016mot16} tracks humans, KITTI~\cite{Geiger2012CVPR, Luiten2020IJCV} tracks pedestrians and cars, and TAO~\cite{dave2020tao} tracks a large taxonomy of 833 categories. In contrast, single object tracking (SOT) follows a single object via a provided initial template of the object, without any detection or recognition. Thus, SOT is often taxonomy-free and operates on generic objects. The community has constructed multiple popular benchmarks to study this problem, including OTB~\cite{wu2013online}, UAV~\cite{mueller2016benchmark}, NfS~\cite{kiani2017need}, TC-128~\cite{liang2015encoding}, NUS-PRO~\cite{li2015nus}, GOT-10k~\cite{huang2019got}, VOT~\cite{VOT_TPAMI}, and TrackingNet~\cite{muller2018trackingnet}.

% MOT involves tracking multiple objects at the same time, which usually requires simultaneously detecting and associating different instances. Example MOT datasets are MOT~\cite{milan2016mot16} for people tracking, KITTI~\cite{Geiger2012CVPR, Luiten2020IJCV} for pedestrian and car tracking, and TAO~\cite{dave2020tao} for tracking moving objects. On the other hand, SOT aims at tracking any objects. In the SOT setting, the tracker is usually given an initial template of the object and is asked to predict a bounding box per frame for the rest of the video sequence. There are many popular benchmarks and datasets for short-term tracking, including OTB~\cite{wu2013online}, UAV~\cite{mueller2016benchmark}, NfS~\cite{kiani2017need}, TC-128~\cite{liang2015encoding}, NUS-PRO~\cite{li2015nus}, GOT-10k~\cite{huang2019got}, VOT~\cite{VOT_TPAMI}, and TrackingNet~\cite{muller2018trackingnet}. Note that these datasets have shorter sequences, with the object is almost always visible.

While these SOT datasets mainly consist of short videos (e.g. a few seconds), long-term tracking has been increasingly of interest. Tracking objects in longer videos (several minutes or more) poses unique challenges, e.g. significant transformations, displacements, disappearances, and reappearances.  On top of localizing the object when visible, the model also must produce no box when the object is absent, and then re-localize the same object when it reappears. OxUvA~\cite{valmadre2018long} is one of the first to benchmark longer videos (average 2 minutes), with 366 evaluation-only videos. LaSOT~\cite{fan2019lasot} scales this to 1400 videos with more frequent object reappearances. Concurrently, VOT-LT~\cite{10.1007/978-3-030-11009-3_1} includes frequent object disappearances and reappearances in 50 purposefully selected videos. 

% \WW{I agree with Kevin, we need to elaborate what's the not in-the-wild property. I thought these videos are carefully picked to have multiple occurrences, and that doesn't really feel problematic. But if in-the-wild actually indicates that the videos are not taken in a normal/ real-world/ daily scenario, then that should be emphasized.}

% Recently, there has been increasing interest in long-term tracking. Long-term tracking not only requires the tracker to predict a bounding box for each frame, but also a confidence score indicating the object presence. Long-term tracking is more challenging as it not only requires the tracker to perform well in short-term tracking, but also to re-detect objects accurately when it reappears. OxUvA~\cite{valmadre2018long} is the first dataset to encourage the thinking of the long-term tracking aspect, by collecting video sequences with an average length of 2 minutes and multiple object absences. However, with only 366 videos, this dataset is relatively small-scale and is annotated sparsely at 1 FPS. LaSOT~\cite{fan2019lasot} is a large-scale densely annotated tracking dataset that not only contains a larger number of sequences (1400), but also includes multiple object disappearance and appearance. VOT-LT~\cite{VOT_TPAMI} is a dataset in the VOT Challenge that is specifically designed to evaluate long-term tracking ability. Although it has a larger number of occurrences per sequence, there are only 50 sequences and not collected in the wild. 

Our \papername\ focuses on long-term SOT and presents multiple critical and unique attributes: 1) significantly larger scale, with \textbf{5708} videos of an average \textbf{6 minutes} (\Cref{fig:scale_comp}); 2) more frequent disappearances \& reappearances (avg. \textbf{17.7} times) happening in natural, real-world scenarios; 3) data sourced from egocentric videos shot in-the-wild, involving unique challenging situations, such as large camera motions, diverse perspective changes, hand-object interactions, and frequent occlusions.

% but with a significant increase in scale compared with previous long-term tracking datasets, as summarized in \Cref{table:long-term_comparison}: a larger number of sequences (17k), longer sequence length (average of 6 minutes per track), and more target disappearances and reappearances (on average 21.5 times per sequence). Moreover, because our videos are sourced from the Ego4D dataset~\cite{grauman2022ego4d}, our dataset exhibit significantly more challenging compared with previous datasets as well, as Ego4D's egocentric in-the-wild data means large camera motion, diverse perspectives, and occlusions, which we believe to more reflective of the challenges of tracking in real-world applications.

\vspace{-2mm}
\subsection{Single object tracking methodologies}
\vspace{-2mm}
% As a classical computer vision problem, a history of attempts have been made to approach SOT. For example, mean-Shift algorithms~\cite{MeanShift} is adapted to tracking objects. Leading tracking systems today typically leverage deep learning 
Many modern approaches use convolutional neural networks (CNNs), either with Siamese network~\cite{li2018high, wang2019fast, li2019siamrpn++} or correlation-filter based~\cite{danelljan2019atom, bhat2019learning, choi2020robust, mayer2021learning, bhat2020know} architectures. With recent successes in vision tasks like classification~\cite{dosovitskiy2020image} and detection~\cite{carion2020end}, Transformer architecture~\cite{vaswani2017attention} for tracking have also become popular. For example, TransT~\cite{chen2021transformer} uses attention-based feature fusion to combine features of the object template and search image. More recently, several works utilize Transformers as direct predictors to achieve a new state of the art, such as STARK~\cite{yan2021learning}, ToMP~\cite{mayer2022transforming} and SBT~\cite{xie2022correlation}. These models tokenize frame features from a ResNet~\cite{he2016deep} encoder, and use a Transformer to predict the bounding box and object presence score with the feature tokens. These methods are often developed on short-term SOT datasets and assume that the target object stays in the field of view with minimum occlusions. On the other hand, long-term trackers~\cite{voigtlaender2020siam,huang2020globaltrack,dai2020high} are designed to cope with the problem of re-detecting objects in their reappearances. Designed to be aware of potential object disappearances, these approaches search the whole image for its reappearance. 

% We carefully benchmark representative methods of these leading tracking systems on the proposed \papername\ dataset and identify their key failures modes. These empirical experiments reveal valuable insights towards promising future directions in egocentric tracking and re-detection. Leveraging these intuitions, we propose multiple simple yet effective changes to the SoTA STARK tracker and train a strong tracker dedicated towards long-term egocentric tracking, EgoSTARK. We hope EgoSTARK can serve as a strong baseline and facilitate future research. 

\begin{table*}[t]
\caption{\textbf{Object tracking datasets comparison.} 
In addition to larger scale than previous datasets, the scenarios captured by EgoTracks represent a significantly harder challenge for SOTA trackers, suggesting room for improved tracking methodology.
% \matt{It isn't clear what story we're telling with columns like 'videos' and 'avg length' -- perhaps we should let table 1 tell the "scale" story and make this table focus on difficulty?  Our overall claim is that we're much larger scale than anything else -- but then we present numbers (like in the 'videos' column) that paint a different picture, which is odd.  I don't think \#videos is a fair way to understand what's going on in the datasets. }
}
\vspace{-2mm}
\centering
\begin{threeparttable}
% \begin{tabular}{llllllll}
\resizebox{0.9\textwidth}{!}{
\centering
\begin{tabular}{cccccccc}
\hline
\textbf{Dataset} & \textbf{\begin{tabular}[c]{@{}c@{}}Video Hours\end{tabular}} & \textbf{Avg. Length (s)} & \textbf{Ann. FPS} & \textbf{Ann. Type}  & \textbf{Egocentric} & \textbf{\begin{tabular}[c]{@{}c@{}}SOTA \\ (P/AO)$^{*}$\end{tabular}} \\ \hline  
ImageNet-Vid ~\cite{ILSVRC15}                                                                       & 15.6           & 10.6           & ~25 & mask & No           &                                 \\
YT-VOS ~\cite{YT-VOS_ECCV}                                                                          & 5.8            & 4.6            & 5            & mask          & No           & -/83.6~\cite{arxiv.2212.14679}  \\
DAVIS 17 ~\cite{Pont-Tuset_arXiv_2017}                                                              & 0.125          & 3              & 24           & mask          & No           & -/86.3~\cite{cheng2022xmem}     \\
TAO~\cite{dave2020tao}                                                                              & 29.7           & 36.8           & 1            & mask          & No           &                                 \\
UVO~\cite{wang2021unidentified}                                                                     & 2.8            & 10             & 30           & mask          & No           & -/73.7~\cite{9008790}           \\
\begin{tabular}[c]{@{}c@{}}EPIC-KITCHENS\\ VISOR\end{tabular}~\cite{darkhalil2022epic} & 36             & 12$^{**}$          & 0.9          & mask          & \textbf{Yes} & -/74.2~\cite{9008790}           \\
GOT-10k ~\cite{huang2019got}                                                                        & 32.8           & 12.2           & 10           & bbox          & No           & -/75.6 ~\cite{cui2022mixformer} \\
OxUvA~\cite{valmadre2018long}                                                                       & 14.4           & 141.2          & 1            & bbox          & No           &                                 \\
{LaSOT~\cite{fan2019lasot}}                                                                         & 31.92          & 82.1           & 30           & bbox          & No           & 80.3/- ~\cite{cui2022mixformer} \\
TrackingNet~\cite{muller2018trackingnet}                                                            & 125.1          & 14.7           & ~28          & bbox          & No           & 86/- ~\cite{cui2022mixformer}   \\
TREK-150~\cite{dunnhofer2021first, dunnhofer2023visual}                                             & 0.45           & 10.81          & 60  & bbox & \textbf{Yes}          & -/50.5 \cite{dunnhofer2021first, dunnhofer2023visual} \\
\textbf{\papername (Ours)}                                                                                       & \textbf{602.9} & \textbf{367.9} & 5            & bbox          & \textbf{Yes} & 45/54.1                        
             
\\ \hline
\end{tabular}
}
% \vspace{-2mm}
\label{table:dataset_overview}
\begin{tablenotes}[para]
     \centering \hspace{-37mm}\footnotesize{$^{*}$: P: Precision, AO: average overlap (J-Score for mask-based datasets). $^{**}$: Original video is 720s.}
\end{tablenotes}
% \footnotesize{*: Original video is 720s. **: An alternative dense set of annotations automatically generated by interpolation is also available.}\\
\end{threeparttable}
\vspace{-3mm}
\end{table*}

% \begin{figure*}[b]
% \vspace{-3mm}
% \hfill
% \begin{subfigure}[b]{0.49\columnwidth}
% \centering
% \includegraphics[width=\columnwidth]{figs/scenario_wordcloud.png}
% % \caption{Word cloud of scenarios and activities.}
% \end{subfigure}
% \hfill
% \begin{subfigure}[b]{0.49\columnwidth}
% \centering
% \includegraphics[width=\columnwidth]{figs/obj_name_wordcloud.png}
% % \caption{Word cloud of object tracks.}
% \end{subfigure}
% \hfill
% \vspace{-1mm}
% \caption{\papername is a large-scale egocentric dataset of diverse scenarios (left) and objects (right).}
% \label{fig:scenario_wordcloud}
% \vspace{-4mm}
% \end{figure*}

\vspace{-2mm}
\subsection{Tracking in egocentric videos}
\vspace{-2mm}
Multiple egocentric video datasets have been introduced in the past decades~\cite{Damen2018EPICKITCHENS,grauman2022ego4d,6247820,10.1007/978-3-319-46454-1_28,Pirsiavash2012DetectingAO,6247805}, offering a host of interesting challenges, many of which require associating objects across frames: activity recognition~\cite{kazakos2019TBN,ego-exo,7410868,Wang2020WhatMT,Ghadiyaram2019LargeScaleWP}, anticipation~\cite{AbuFarha2018WhenWY,FurnariLSTM20,9710253}, video summarization~\cite{7750564,6247820,10.1007/s11263-014-0794-5,6619194}, human-object interaction~\cite{darkhalil2022epic,Liu2019ForecastingHO}, episodic memory~\cite{grauman2022ego4d}, visual query~\cite{grauman2022ego4d}, and camera-wearer pose inference~\cite{8099856}. To tackle these challenges, tracking is leveraged in many methodologies~\cite{grauman2022ego4d,darkhalil2022epic,7293645,10.1007/s11263-014-0794-5,Liu2019ForecastingHO}, yet few works have been dedicated to this fundamental problem on its own. \cite{dunnhofer2021first, dunnhofer2023visual} have started to recognize the challenges of egocentric object tracking and might be the most related work to ours. The major difference, however, is the scale of the dataset: \cite{dunnhofer2021first, dunnhofer2023visual} contain 150 tracks intended only for evaluation, while \papername is 100$\times$ larger (see \Cref{table:dataset_overview}), containing 20k tracks with training and evaluation splits. Also, while past efforts have sourced videos from the kitchen-heavy EPIC-KITCHEN \cite{Damen2018EPICKITCHENS}, \papername sources videos from Ego4D \cite{grauman2022ego4d}, which has more diverse scenarios.
% The ones that do have started to recognize the challenges of egocentric object tracking \cite{dunnhofer2021first, dunnhofer2023visual}, though at smaller scales.
\papername provides a unique, large-scale testbed for developing tracking methods dedicated to egocentric videos; our improved baseline EgoSTARK also serves as a potential plug-and-play module to solve other tasks where object association is desired. 

In egocentric video understanding, Ego4D~\cite{grauman2022ego4d} and EPIC-KITCHENS VISOR~\cite{darkhalil2022epic} are closely related. Ego4D contains the largest collection of egocentric videos in-the-wild; \papername is annotated on a subset of Ego4D. In addition, Ego4D proposes many novel tasks, such as Episodic Memory, with tracking identified as a core component. VISOR was introduced concurrently, annotating short-term (12 sec on average) videos from EPIC-KITCHENS~\cite{Damen2018EPICKITCHENS} with instance segmentation masks. We believe \papername offers multiple unique values complementary to EPIC-VISOR: long-term tracking (6 min vs. 12 sec), significantly larger-scale (5708 video clips vs. 158), and more diversified video sources (80+ scenes vs. kitchen-only; see 
Appendix A).

\vspace{-2mm}
\section{The \papername ~dataset}
\vspace{-2mm}
We present \papername: a large-scale long-term egocentric single object tracking dataset, consisting of a total of 22028 tracks from 5708 videos. We follow the same data split as the Ego4D Visual Queries (VQ) 2D benchmark.(See Supplementary for details).

% videos for training, 1.2k for validation, and 1.1k reserved as a held-out test set for the challenge. Comparisons to other datasets are provided in \Cref{table:long-term_comparison,table:dataset_overview}.

% \subsection{From Ego4D Visual Queries to \papername}
\vspace{-2.5mm}
\subsection{Ego4D visual queries (VQ) benchmark} 
\label{sec:ego4d_vq}
\vspace{-2.5mm}
Ego4D~\cite{grauman2022ego4d} is a massive-scale egocentric video dataset, consisting of 3670 hours of diverse daily-life activities of consenting participants in an in-the-wild format; the videos have personally identifiable information removed and were screened for offensive content. The dataset is accompanied by multiple benchmarks, 
% such as episodic memory, hands and objects, social interaction, and forecasting. 
but the most relevant task for our purposes is episodic memory's 2D VQ task: Given an egocentric video and a cropped image of an object, the goal is to localize when and where the object was last seen in the video, as a series of 2D bounding boxes in consecutive frames. 
% Such a task can be used to test for example an augmented reality (AR) assistant's ability to answer questions such as ``Where is [picture of my mug]?'', which requires remembering objects that the user might be interested in and localizing them in video. 
This task is closely related to long-term tracking: finding an object in a video given a visual template is identical to the re-detection problem in long-term tracking. 
%Moreover, baselines used in this task is closely tied to tracking methods: it consists of running a Siam-RCNN global tracker and a KYS local tracker to output the required response track. 
Moreover, Ego4D's baselines rely heavily on tracking methods: Siam-RCNN~\cite{voigtlaender2020siam} and KYS~\cite{bhat2020know} for global and local tracking, respectively.

\noindent \textbf{Shortcomings.}
While highly related, the VQ dataset is not immediately suitable long-term tracking. 
In particular, the VQ annotation guidelines were roughly the following: 1) identify three different objects that appear multiple times in the video; 2) annotate a query template for each object, which should contain the entire object without any motion blur; 3) annotate an occurrence of the object that is temporally distant from the template.
Thus, these annotations are not exhaustive over time (they are quite sparse), limiting their applicability to tracking.
On the other hand, the selection criteria result in a strong set of candidate objects, which we leverage to build \papername.

\vspace{-2.5mm}
\subsection{Annotating VQ for long-term tracking} 
\vspace{-2.5mm}
We thus start with the VQ visual crop and response track, asking annotators to first identify the object represented by the visual crop, the response track, and object name. From the video's start, we instruct the annotators to draw a bounding box around the object each time it appears. Because annotators must go through each video in its entirety, which contain an average of $\sim$1800 frames at 5 frames per second (FPS), this annotation task is labor-intensive, taking roughly 1 to 2 hours per track.  An important aspect of this annotation is its exhaustiveness: the entire video is densely annotated for the target object, and any frame without a bounding box is considered as a negative. Being able to reject negatives examples is an important component to re-detection in real-world settings, as false positives can impact certain applications as much as false negatives. 

% In \papername\, given the visual crop as the template for the target object, the tracker is required to start from the beginning of video and report a bounding box each time it sees it. 

% By extending the 2D VQ task from the Ego4D dataset and benchmark suite, we provide a densely annotated long-term object tracking dataset that can be used to comprehensively evaluate the re-detection ability of the trackers. The entire annotation is split into three sets: train, validation, and test. We plan to make the train and validation sets publicly available to train and evaluate long-term trackers, while holding the test set private for public challenges. 

\noindent \textbf{Quality Assurance.} All tracks are quality checked by expert annotators after the initial annotations. To measure the annotation quality, we employ multi-review on a subset of the validation set. Three independent reviewers are asked to annotate the same video.  We find the overlaps between these independent annotations are high ($>0.88$ IoU). Further, since \papername has a focus on re-detection, we check the temporal overlap of object presence and find it to be consistent across annotators. 
In total, the entire annotation effort represented roughly 86k worker-hours of effort.
% \WW{If we run out of space, we can mention one sentence and move it to supp.}

\begin{figure}[t]
\begin{minipage}[c]{0.49\linewidth}
\includegraphics[width=\linewidth]{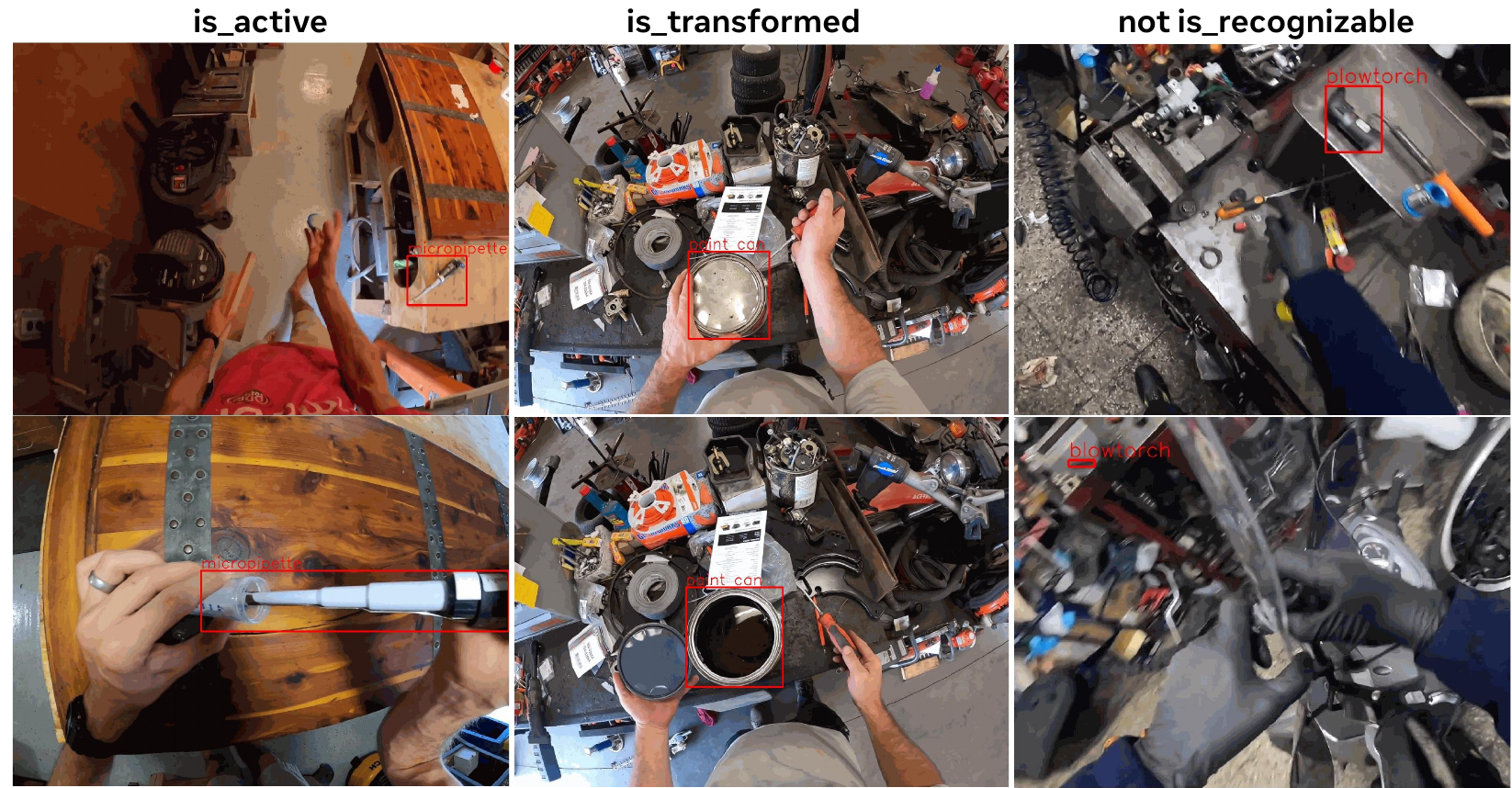}
\caption{\textbf{\papername tracklet attribute examples.} \textit{Left}: Micropipette on a bench (top) versus actively used (bottom). \textit{Middle}: A paint can (top) is opened (bottom). \textit{Right}: A blowtorch (top) requiring context from other frames to identify due to distance and motion blur (bottom). 
% A hard to recognize blowtorch (bottom) due to distance and motion blur; annotators must rely on context from other frames to identify the object. }
}
\vspace{-5mm}
\label{fig:attribute_vis}
\end{minipage}
\hfill
\begin{minipage}[c]{0.49\linewidth}
\centering
% \vspace{-3mm}

\captionof{table}{\textbf{Track attributes} in train/val sets.}
% \vspace{-3mm}
\resizebox{0.85\columnwidth}{!}{
\begin{tabular}{lcc}
\hline
                               & \textbf{Total number} & \textbf{Percentage} \\ \hline
All Tracks                & 17593                  & 100\%               \\
\texttt{is\_active}            & 3963                  & 22.52\%             \\
% \texttt{is\_location\_changed} & 942                   & 21.12\%             \\
\texttt{is\_transformed}       & 1080                   & 6.13\%              \\
\texttt{is\_recognizable}      & 17557                  & 99.79\%             \\ \hline
\end{tabular}
}
\label{table:occ_attribute}

\vspace{3mm}

\includegraphics[width=0.9\linewidth]{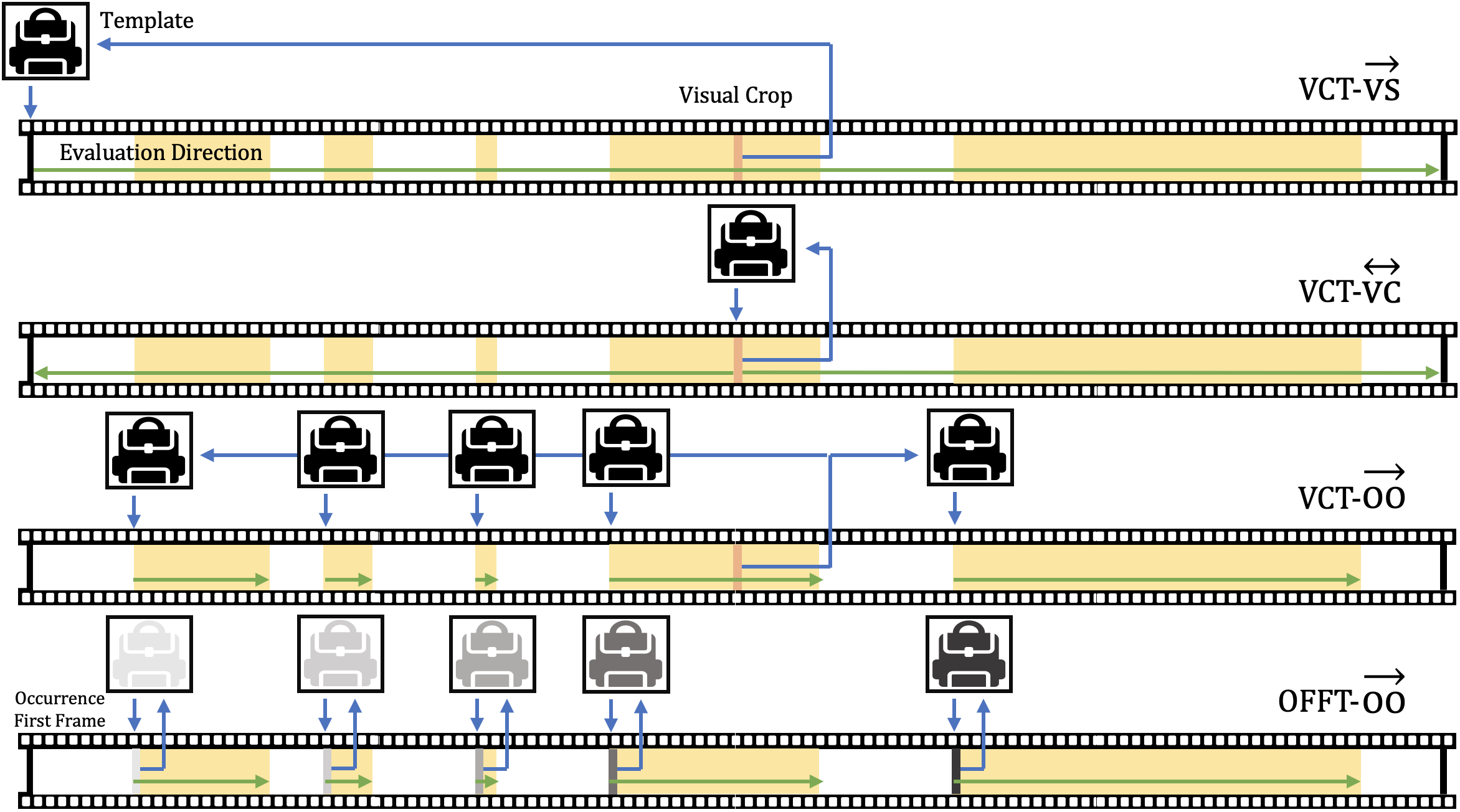}
\caption{Evaluation protocols visualization.}
\label{fig:eval_proto}
\vspace{-10mm}
\end{minipage}%
\end{figure}

\vspace{-2.5mm}
\subsection{Tracklet attributes}
\label{sec:tracklet_attributes}
\vspace{-2.5mm}
In addition to the bounding box annotations, we also label certain relevant attributes to allow for different training strategies or deeper analysis of validation set performance. 
% One object track contains many occurrences and each attribute is attached to each occurrence of the object. 
We annotate the following three attributes per occurrence (see \Cref{fig:attribute_vis} for examples and \Cref{table:occ_attribute} for statistics): 
% is\_active, is\_transformed, is\_moved, is\_recognizable.
\begin{tight_itemize}
\item \texttt{is\_active}: In Ego4D, the camera wearer often interacts with relevant objects with their hands. Objects in the state of being handled pose a challenge for tracking algorithms due to frequent occlusion and rapid changes in pose.
\item \texttt{is\_transformed}: Objects in Ego4D may undergo transformations, such as deformations and state changes. Such instances require being able to quickly adapt to the tracked object having a new appearance.
% \item \texttt{is\_location\_changed}: Some objects in \papername change location off camera; we indicate if an object has changed locations in an occurrence compared to its previous occurrence. To re-detect such objects, a model cannot overfit to the object's context. 
\item \texttt{is\_recognizable}: Due to occlusions, motion blur, scale, or other conditions, some objects in Ego4D can be extremely difficult to recognize without additional context. We thus annotate if the object is recognizable solely based on its appearance, without using additional context information (\eg other frames). 
\end{tight_itemize}

\begin{figure}[t]
\includegraphics[width=\columnwidth]{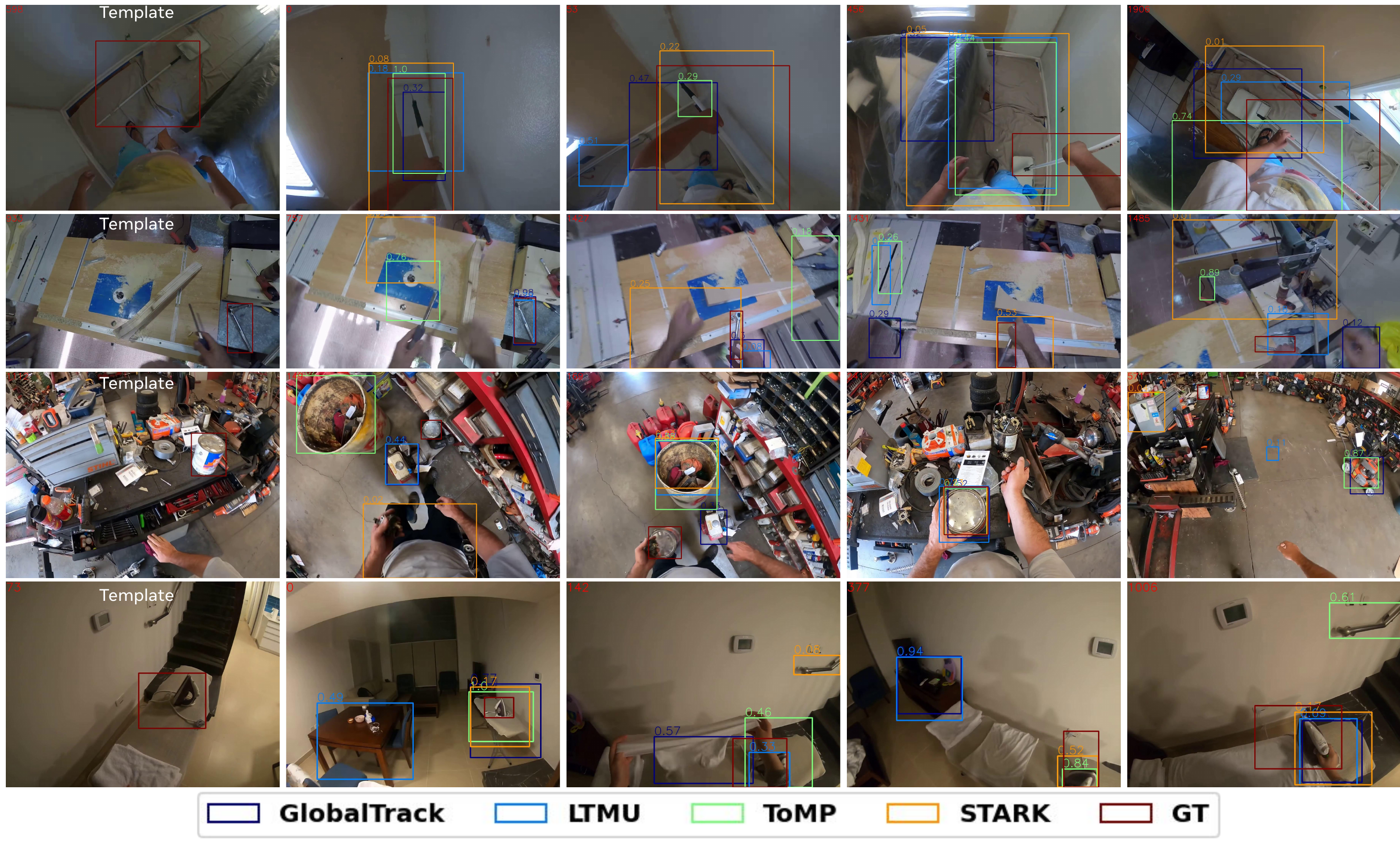}
\caption{{Qualitative results of different trackers. EgoTracks presents significant challenges for all trackers, due to drastic viewpoint changes, occlusions, changes in scale, head motion etc.} }
\vspace{-5mm}
\label{fig:comparison_trackers}
\end{figure}

\vspace{-2mm}
\section{Analysis of state-of-the-art SOT trackers} \label{sec:sota_exp}
\vspace{-2mm}
We compare the performance of several off-the-shelf tracking models on \papername's validation set. 
Identifying STARK~\cite{yan2021learning} as the one with the best performance, 
% and propose several training and test-time tricks to train an improved STARK model which we call Ego-STARK. 
we conduct further ablation studies under different settings using STARK to further understand its behavior.

\vspace{-2mm}
\subsection{Evaluation protocols and metrics}
\vspace{-2mm}
\noindent \textbf{Evaluation Protocols.} We introduce several evaluation protocols for \papername, consisting of different combinations of the initial template, evaluated frames, and the temporal direction in which the tracker is run. For the initial template, we consider two choices:
\begin{tight_itemize}
\item \textbf{Visual Crop Template (VCT)}: The visual crop images were specifically chosen to be high-quality views of the target and served as our annotators' references for identifying the object throughout the videos. Thus, they make ideal candidates for initializing a tracker.
\item \textbf{Occurrence First Frame Template (OFFT)}: The tracker is initialized with the first frame of each occurrence (see $\overrightarrow{\text{OO}}$ below). While this may result in a lower quality view of the object, temporal proximity to subsequent frames means it may be closer in appearance.
\end{tight_itemize}
Note that we exclude the template frame from the calculation of any evaluation metrics. 
We also consider several choices for the evaluated frames and temporal direction:
\begin{tight_itemize}
\item \textbf{Video Start Forward ($\overrightarrow{\text{VS}}$)}: The tracker is evaluated on every frame of the video in causal order, starting from the first frame. This represents a tracker's ability to follow an object through a long video.
\item \textbf{Visual Crop Forward/Backward ($\overleftrightarrow{\text{VC}}$)}: The tracker is run on the video twice, once starting at the visual crop frame and running forward and time, and a second time running backwards. 
This represents an alternative way of covering every frame in the video, but with closer visual similarity between \textbf{VCT} initialization and the first frames encountered by the tracker.
\item \textbf{Occurrences Only Forward ($\overrightarrow{\text{OO}}$)}: The tracker is only evaluated on the object occurrences, when the object is visible. This simplifies the tracking task and allows us to dis-entangle the challenge of re-detection from that of simply tracking in an egocentric clip.
\end{tight_itemize}
We specify protocols by concatenating the appropriate descriptors. 
We primarily consider \textbf{VCT-$\overrightarrow{\text{VS}}$}, \textbf{VCT-$\overleftrightarrow{\text{VC}}$}, \textbf{VCT-$\overrightarrow{\text{OO}}$}, and \textbf{OFFT-$\overrightarrow{\text{OO}}$} (\cref{fig:eval_proto}) in our experiments.

\noindent \textbf{Metrics.} 
We adopt common metrics in object tracking, including F-score, precision, and recall; details can be found in~\cite{lukevzivc2018now}.
Trackers are ranked mainly by the F-score.
% The final metrics used for ranking trackers are mainly based on the F-score.
We additionally consider average overlap (AO), success, precision, and normalized precision as short-term tracking metrics~\cite{tao2017tracking}.

\vspace{-2mm}
% \subsection{Comparison of SOT trackers}
\subsection{SOT trackers struggle on EgoTracks}
\vspace{-2mm}
\begin{table}[t]
\centering
\caption{\textbf{\papername performance comparison.} Off-the-shelf, all trackers perform poorly, demonstrating the new challenges of \papername.
Higher performance from tracking by detection methods + Oracle imply that instance association, not detection, is one of the primary challenges.
}
% \vspace{-2mm}
\resizebox{0.7\columnwidth}{!}{
\begin{tabular}{lccccc}
\hline
{ \textbf{Method}} & { \textbf{AO}} & { \textbf{F-score}} & { \textbf{Precision}} & { \textbf{Recall}} & \textbf{FPS} \\ \hline
{ KYS~\cite{bhat2020know}}             & { 16.09}       & { 13.09}       & { 12.50}              & { 13.74}     & 20      \\ 
{ DiMP~\cite{bhat2019learning}}            & { 16.45}       & { 11.84}       & { 10.31}              & { 13.91}       & 43    \\ 
{ GlobalTrack~\cite{huang2020globaltrack} $^{\dag}$}     & { 23.63}       & { 20.40}       & { 31.28}              & { 15.14}        & 6   \\ 
{ LTMU~\cite{dai2020high} $^{\dag}$}      & { 29.33}       & { 27.46}       & { 37.28}              & { 21.74}    & 13       \\ 
{ ToMP~\cite{mayer2022transforming}}            & { 30.93}       & { 20.95}       & { 19.63}              & { 22.46}       & 24.8    \\ 
Siam-RCNN \cite{voigtlaender2020siam} $^{\dag}$     & 37.48       & 35.38       & 52.80              & 26.67 & 4.7 \\
MixFormer (MixViT-L, ConvMAE) \cite{cui2022mixformer,cui2023mixformer}       & 27.93       & 25.54       & 28.30              & 23.27           & 10           \\

{ STARK~\cite{yan2021learning} - Res50}   & { 35.99}       & { 30.48}       & { 34.70}              & { 27.17}    & 41.8       \\ 
{ STARK~\cite{yan2021learning} - Res101}  & { 35.03}       & { 30.18}       & { 35.30}              & { 26.35}   & 31.7        \\ \hline
\hline
\multicolumn{5}{l}{\textbf{Tracking by Detection} $^{\dag}$} \\
Mask R-CNN~\cite{he2017mask}+Oracle & 60.00 & - & - & - \\
GGN~\cite{wang2022open}+Oracle & 75.92 & - & - & - \\
% GGN+ImageNet & 2.50 & 1.22 & 0.8 & 2.5 \\
GGN+InstEmb & 15.19 & 9.92 & 11.75 & 8.58 \\
\hline
\end{tabular}
}
\begin{tablenotes}[para]
     \centering \hspace{-90mm}\footnotesize{{$^{\dag}$: trackers with re-detection}}
\end{tablenotes}
\vspace{-5mm}
\label{table:sota_comparison}
\end{table}
% All trackers use the visual crop as the initial object template and then are asked to track the object starting from the first frame of each video, which we refer to as protocol 1 (\textbf{proto1}). The visual crop is carefully selected in Ego4D because the annotators are instructed to choose the frame where it contains the entire object and the object is clearly visible with no motion blur and occlusion, making it ideal as the template.

We compare the performance of select trackers on \papername with the \textbf{VCT-$\overrightarrow{\text{VS}}$} evaluation protocol. Given the breadth of tracking algorithms, we do not aim to be exhaustive but select high-performing representatives of different tracking principles. KYS~\cite{bhat2020know} and DiMP~\cite{bhat2019learning} are two short-term tracking algorithms that maintain an online target representation. ToMP~\cite{mayer2022transforming}, STARK~\cite{yan2021learning} and MixFormer~\cite{cui2022mixformer, cui2023mixformer} are three examples of the SOTA short-term trackers based on Transformers.
% that use transformer-based architectures.
GlobalTrack~\cite{huang2020globaltrack} is a global tracker that searches the entire search image for re-detection. LTMU~\cite{dai2020high} is a high performance long-term tracker that combines a global tracker (GlobalTrack) with a local tracker (DiMP). Siam R-CNN~\cite{voigtlaender2020siam} leverages dynamic programming to model a full path of history for long-term.
% There are many tracking algorithms and the trackers we choose are not meant to be exhaustive but a representative set for different tracking principles. 
The performance of these trackers on \papername are summarized in \Cref{table:sota_comparison}. AO in this table is equivalent to recall at the probability threshold 0. Qualitative results are shown in \Cref{fig:comparison_trackers}.

We highlight several observations. First, the object presence scores from most short-term trackers are not very useful, as can be seen from the low precision of KYS (12.5), DiMP (13.91), and ToMP (19.63), while long-term trackers like GlobalTrack, DiMP\_LTMU and Siam R-CNN achieve higher precisions at 31.28, 37.28 and 52.8. This is expected as long-term trackers are designed to place more emphasis on high re-detection accuracy, though there clearly is still room for improvement. STARK achieves the second highest precision at 34.70, which is an exception as it has a second training stage to teach the model to classify whether the object is present. Second, more recent works such as MixFormer and STARK achieve better F-score than previous short-term trackers. This could be partially due to advances in training strategies, more data, and Transformer-based architectures. Surprisingly, we found recent MixFormer~\cite{cui2023mixformer} does not outperform STARK, despite achieving new SOTA on its training dataset. This highlights a potential difficulty in generalization.

We also include results using the principle of Tracking by Detection~\cite{boostedPFOkuma04,1467482}: a detector proposes 100 bounding boxes, and we select the best using cosine similarity of box features. We observe that an open-world detector GGN~\cite{wang2022open} trained on COCO~\cite{lin2014microsoft} generalize reasonably well with oracle matching, achieving 75.92 AO. However, the association problem is very challenging, bringing down the AO to 15.19. Implementation details can be found in Appendix B.

\vspace{-2mm}
% \subsection{Alternative evaluation protocols}
\subsection{Re-detection and diverse views are challenging}
\vspace{-2mm}
\begin{table}[t]
\centering
%\caption{Comparison of different tracker initializations. The upper table shows different trackers initialized from the initial frame in each occurrence and only asked to track each occurrence (no re-detection or perfect re-detection). The lower table shows STARK's performance when asked to track starting from different frames.}
\caption{Comparing tracker initializations. (Left) Comparison of trackers initialized from the first frame in each occurrence and tracking only that single occurrence (oracle re-detection). (Right) STARK whole-video performance, starting from video start frame vs.~the visual crop frame.}
% \vspace{-2mm}
\parbox{.45\linewidth}{
    \centering
\resizebox{\linewidth}{!}{
    \begin{tabular}{lcccc}
    \hline
    \textbf{Method} & \textbf{AO} & \textbf{Success} & $\mathbf{Pre}$ & $\mathbf{Pre_{norm}}$\\ \hline
    % KYS \cite{bhat2020know}            & 34.87            & 31.22              & 34.87   \\   
    % DiMP \cite{bhat2019learning}          & 35.70            & 32.13              & 38.98                \\
    % ToMP \cite{mayer2022transforming}           & 45.93            & 41.74              & 47.88                \\
    % STARK \cite{yan2021learning}          & 50.64            & 45.76              & 51.91                \\ 
    % STARK \cite{yan2021learning} - \textbf{VQI}           &             &              &                 \\ \hline 
    
    KYS \cite{bhat2020know}             & 33.92       & 34.87            & 31.22              & 34.87                \\
    DiMP \cite{bhat2019learning}           & 34.80       & 35.70            & 32.13              & 38.98                \\
    ToMP \cite{mayer2022transforming}            & 45.17       & 45.93            & 41.74              & 47.88                \\
    STARK \cite{yan2021learning}           & 50.01       & 50.64            & 45.76              & 51.91       \\ \hline 
    % \hline
    % STARK - \textbf{OFFT-$\overrightarrow{\text{OO}}$}          & 48.18       & 48.86            & 45.76              & 51.90                \\
    % STARK - \textbf{VCT-$\overrightarrow{\text{OO}}$} & 51.10       & 51.83            & 47.78              & 53.96   \\ \hline
    \end{tabular}
}
}
\hfill
% \vspace{-3mm}
\parbox{.52\linewidth}{
\resizebox{\linewidth}{!}{

    \begin{tabular}{lcccc}
    \\ \hline
    \textbf{Method}                          & \textbf{AO} & \textbf{F-score} & \textbf{Precision} & \textbf{Recall} \\ \hline
    STARK - \textbf{VCT-$\overrightarrow{\text{VS}}$} & 35.99       & 30.48       & 34.70              & 27.17           \\
    STARK - \textbf{VCT-$\overleftrightarrow{\text{VC}}$} & 40.01       & 34.02       & 38.31              & 30.60    \\ \hline      
    \end{tabular}
}
}
\vspace{-5mm}
\label{table:occ_comparison}
\end{table}

% The previous comparison requires trackers to rank their predictions based on object presence confidence, which might not be fair for short-term trackers who were trained mainly to predict accurate bounding boxes. 
% Thus, here we show results using alternative tracking protocols: 1) The tracker is initialized using the first frame of each occurrence and then track consecutively until the end of the occurrence (\textbf{$\overrightarrow{\text{FFI}}$}); 2) The tracker is initialized using the visual template, but instead of running from the first frame of the video, the tracker runs forward from the visual crop until the end and backward until the beginning - (\textbf{$\overleftrightarrow{\text{VQI}}$}). 
% For 1), we re-run all previous short-term tracker, while for 2) we only experiment with STARK. By default, we use STARK with Res50 backbone. We show in \Cref{table:occ_comparison} the performance comparison in terms of success, precision and normalized precision. As we can see, performance of these short-term trackers is much higher than \Cref{table:sota_comparison}. SoTA trackers perform reasonably well on data from a different domain. Even though there is still a significant drop in success and precision, the drop is smaller than that of the long-term tracking protocol. This further emphasizes the challenges in tracking object long-term. 

We perform additional EgoTracks experiments following alternative evaluation protocols to gain further insights on tracker performance (\Cref{table:occ_comparison}).
% First, one of \papername's strengths is that it brings the challenge of re-detection to the forefront, with far more disappearances and reappearances than previous datasets (\Cref{table:long-term_comparison}). 
% This is especially hard on short-term trackers, which are mostly designed to output a bounding box at every frame.
To decouple the re-detection problem from  other egocentric aspects of \papername, 
% To understand how much of the challenge of \papername is from the re-detection problem, 
we evaluate with the \textbf{OFFT-$\overrightarrow{\text{OO}}$} protocol, which ignores the negative frames of the video, thus obviating the need for re-detection.
Unsurprisingly, all trackers do significantly better, emphasizing the challenging nature of re-detection in \papername.
We also run experiments in the \textbf{VCT-$\overleftrightarrow{\text{VC}}$} setting, where the initial template is temporally adjacent to the first tracked frames.
Here we see a 3-4\% improvement to AO, F-score, precision, and recall compared to the \textbf{VCT-$\overrightarrow{\text{VS}}$} protocol, illustrating that trackers like STARK are designed to expect gradual transitions in appearance.
Both these experiments illustrate that the re-detection problem is a significant challenge for tracking and the need for better long-term benchmarks.

\vspace{-2mm}
% \subsection{Effect of attributes on tracking performance}
\subsection{Attributes capture hard scenarios for tracking}
\vspace{-2mm}
% In \Cref{sec:tracklet_attributes}, we describe additional tracklet attribute annotations for the validation set.
We use the validation set tracklet attribute annotations described in \Cref{sec:tracklet_attributes} to further understand performance on our evaluation set.
For each attribute, we split the tracklets into two groups, corresponding to the attribute being true and false. 
We then use a standard STARK tracker~\cite{yan2021learning} and report AO for each group of tracklets using the \textbf{OFFT-$\overrightarrow{\text{OO}}$} evaluation protocol in \Cref{table:attributes_performance}.
As might be expected, we find that when objects are being actively used by the user or in the midst of a transformation, AO tends to be lower, by roughly 6\%, likely due to occlusions or changes in appearance.
Additionally, STARK tends to have a harder time when the object is hard to recognize in the image, whether due to occlusions, blur, scale, or other conditions.

\begin{table}[t]
    \parbox{.40\linewidth}{
        \centering
        \caption{\textbf{OFFT-$\protect\overrightarrow{\text{OO}}$} AO of standard STARK model~\cite{yan2021learning} for each attribute.}
        % , averaged across tracklets.}
        \label{table:attributes_performance}
        \resizebox{0.39\columnwidth}{!}{
        \begin{tabular}{lcc}
        \hline
        \textbf{Attribute}        & \textbf{True} & \textbf{False} \\ \hline
        \texttt{is\_active}       & 49.65         & 55.73          \\
        \texttt{is\_transformed}  & 49.19         & 55.31          \\
        \texttt{is\_recognizable} & 55.52         & 46.65          \\ \hline
        \end{tabular}
        }
    }
    \hfill
    \parbox{.57\linewidth}{
        \centering
        \caption{Performance of trackers finetuned on \papername.}
        \label{table:finetuned_comparison}
        \resizebox{0.56\columnwidth}{!}{
        \begin{tabular}{lccccc}
        \hline
        \textbf{Method} & \textbf{AO} & \textbf{F-score} & \textbf{Precision} & \textbf{Recall} \\ \hline
        % KYS             &             &             &                    &                 \\
        % Dimp            &             &             &                    &                 \\
        ToMP            & 36.13       & 28.11       & 29.01              & 27.26           \\
        Siam-RCNN       & 45.67       & 41.41       & 56.11              & 32.81           \\
        STARK           & 44.25       & 38.20       & 42.06              & 34.99         \\
        \hline
        \end{tabular}
        }
    }
    \vspace{-3mm}
\end{table}

% we annotate three attributes for each tracklet in each track. Tracklets with different attributes exhibit different level of difficulty for tracking. For example, object in
% \begin{table}
% \caption{\textbf{OFFT-$\protect\overrightarrow{\text{OO}}$} AO of standard STARK model~\cite{yan2021learning} for each attribute, averaged across tracklets.}
% % \vspace{-2mm}
% \centering
% % \resizebox{0.65\columnwidth}{!}{
% \begin{tabular}{lcc}
% \hline
% \textbf{Attribute}        & \textbf{True} & \textbf{False} \\ \hline
% \texttt{is\_active}       & 49.65         & 55.73          \\
% \texttt{is\_transformed}  & 49.19         & 55.31          \\
% \texttt{is\_recognizable} & 55.52         & 46.65          \\ \hline
% \end{tabular}
% % }
% \vspace{-2mm}
% \label{table:attributes_performance}
% \end{table}

\vspace{-2mm}
% \section{Ego-STARK}
\section{Egocentric tracking design considerations}
\label{sec:ego_design}
\vspace{-2mm}
% With STARK being the most competitive among other trackers, we adopt it as the main algorithm we use as the base for improvement and for ablation study. 
% Although STARK does not specifically aim at addressing the long-term tracking problem, it trains the model in two stages which makes it suitable to predict object presence. In the first stage, the model is trained to only predict bounding box by sampling frames where the object is visible. In the second stage, only a few extra layers for the classification head are finetuned, by sampling 50\% of the time a positive example where the object is visible and 50\% of time a negative example where the object is not visible. 

Observing that existing trackers do not perform well on EgoTracks, we perform a systematic exploration of priors and other design choices for egocentric tracking.
Though not specifically designed for long-term tracking, \Cref{sec:sota_exp} suggests STARK~\cite{yan2021learning} to be the most competitive tracker on \papername. We focus on this tracker for additional analysis, suggesting improvements to egocentric performance.

% STARK model summary here? (Not sure how necessary it is?)

\vspace{-2mm}
% \subsubsection{Training and testing tricks}
% \subsection{Dataset finetuning}
\subsection{Egocentric finetuning is essential}
\vspace{-2mm}

\begin{table}[t]
    \vspace{-4mm}
    \parbox{.54\linewidth}{
        \centering
        \caption{Train/test-time hyperparameters comparison.}
        \label{table:stark_tricks_comparison}
        \resizebox{0.54\columnwidth}{!}{
        \begin{tabular}{l|lcccc}
        \hline
        \textbf{} & \textbf{Method} & \textbf{AO} & \textbf{F-score} & \textbf{Precision} & \textbf{Recall} \\ \hline
        \multirow{3}{*}{\textbf{Data}}       & STARK                            & 35.99    & 30.48       & 34.70              & 27.17           \\
        & STARK - ft on VQ          & 38.94       & 33.53       & 39.13              & 29.33           \\
        & STARK - ft on \papername & 44.25       & 38.20       & 42.06              & 34.99    \\ \hline
        \multirow{2}{*}{\textbf{Augmentation}}   & STARK - ft on VQ          & 38.94       & 33.53       & 39.13              & 29.33           \\
        & STARK - ft + multiscale   & 48.44       & 41.92       & 42.65              & 41.30           \\ \hline
        \multirow{4}{*}{\textbf{Search window}} & search\_size = 320               & 35.99       & 30.48       & 34.70              & 27.17           \\
        & search\_size = 480               & 48.21       & 39.69       & 43.95              & 36.19           \\
        & search\_size = 640               & 52.09       & 42.39       & 46.23              & 39.15           \\
        & search\_size = 800               & 54.08       & 43.74       & 47.60              & 40.45 \\ \hline          
        \end{tabular}
        }
    }
    \vspace{-3mm}
    \hfill
    \parbox{.44\linewidth}{
        \small
        \centering
        \caption{STARK with different context ratios. Bold row is the default setting. \textbf{CR}: context ratio, \textbf{SRR}: search region ratio, \textbf{SIS}: search image size (in image resolution).}
        \label{table:context_ratio}
        \resizebox{0.44\columnwidth}{!}{
        \begin{tabular}{llll|cccc}
        \hline
        \multicolumn{4}{c|}{\textbf{Method}}                                               & \multirow{2}{*}{\textbf{AO}} & \multirow{2}{*}{\textbf{F-score}} & \multirow{2}{*}{\textbf{Precision}} & \multirow{2}{*}{\textbf{Recall}} \\ 
        \textbf{Setting}                                   & \textbf{CR} & \textbf{SRR} & \textbf{SIS} &                              &                              &                                     &                                  \\ \hline
        \multirow{4}{*}{\textbf{Same SIS}}   & 1x                       & 2.5x                           & 320                        & 28.22                        & 26.81                        & 28.68                               & 25.16                            \\
        & \textbf{2x}              & \textbf{5x}                    & \textbf{320}               & 38.94                        & 33.53                        & 39.13                               & 29.33                            \\
        & 3x                       & 7.5x                           & 320                        & 44.70                        & 36.03                        & 40.28                               & 32.59                            \\
        & 4x                       & 10x                            & 320                        & 43.19                        & 34.32                        & 37.98                               & 31.31                            \\ \hline
        \multirow{2}{*}{\textbf{Same SRR}} & 1x                       & 5x                             & 640                        & 41.50                        & 31.09                        & 30.31                               & 31.91                            \\
        & 3x                       & 5x                             & 208                        & 39.87                        & 35.36                        & 41.54                               & 30.79                            \\ \hline
        \multirow{3}{*}{\textbf{Same CR}}       & 2x                       & 7.5x                           & 480                        & 48.21                        & 39.69                        & 43.95                               & 36.19                            \\
        & 2x                       & 10x                            & 640                        & 52.09                        & 42.39                        & 46.23                               & 39.15                            \\
        & 2x                       & 12.5x                          & 800                        & 54.08                        & 43.74                        & 47.60                               & 40.45     \\ \hline                    
        \end{tabular}
        }
    }
    \vspace{-3mm}
\end{table}

We first demonstrate how various trackers trained on third-person videos can significantly benefit from finetuning on EgoTracks. As shown in \Cref{table:finetuned_comparison}, all methods gain improvement on F-score ranging from 6\% - 10\%. In addition, as shown in \Cref{table:stark_tricks_comparison}, finetuning on the VQ response track subset improves the F-score from 30.48\% to 33.53\%, while using the full \papername annotation further improves the F-score by 4.67\% to 38.2\%. This demonstrates that: 1) finetuning with egocentric data helps close the exocentric-egocentric domain gap; 2) training on 
full \papername provides further gains, showing the value of our training set.

% We first demonstrate how STARK trained on third-person videos significantly benefits from finetuning on egocentric data.
% % analyzing the impact of training data, as well as how the characteristics of the data impact the model's assumptions.
% We experiment with two versions of \papername: the Ego4D VQ response track dataset (i.e. short-term subset of \papername) and the full \papername training set, which contains more point-of-view and scale variations, as well as hard negatives.
% % We do this to demonstrate the importance of having a more diverse dataset (more point-of-view and scale changes). 
% As shown in \Cref{table:stark_tricks_comparison}, finetuning on the VQ response track subset improves the F-score from 30.48\% to 33.53\%. 
% Using the full \papername annotation further improves the F-score by 4.67\% to 38.2\%. 
% This demonstrates that: 1) finetuning with egocentric data helps close the domain gap; 2) training on full \papername provides further gains, showing the value of our training set.

\vspace{-2mm}
% \subsection{Adjusting spatiotemporal priors}
\subsection{Third-person spatiotemporal priors fail}
\label{sec:spacetimepriors}
\vspace{-2mm}
Modern SOTs find certain assumptions on object motion, appearance, and surroundings helpful on past datasets, but some of these design choices translate poorly to long-term egocentric videos.

\noindent \textbf{Search window size.} 
An example is local search. Many trackers assume the tracked object appears within a certain range of its previous location. 
Thus, for efficiency, these methods often search within a local window of the next frame. 
This is reasonable in high FPS, smooth videos with relatively slow motion, commonly in previous short-term tracking data, but in egocentric videos, the object's pixel coordinates can change rapidly (frequent large head motions), and re-detection becomes a key problem. 
Therefore, we experiment with expanded search regions beyond what are common in past methods.
As we expand search size from 320 to 800, we see dramatic improvements (\Cref{table:stark_tricks_comparison}): STARK is able to locate objects that were previously outside search window due to rapid movements.

\noindent \textbf{Multiscale augmentations.} The characteristics of egocentric video also affect common SOT assumptions of object scale. 
Many trackers are trained with the assumption that an object's scale is consistent with the template image and between adjacent frames.
% , thus affecting both search regions and learned biases in bounding box prediction.
However, large egocentric camera motions, locomotion, and hand interactions with objects (\eg bringing an object to one's face, as in eating) can translate to objects rapidly undergoing large changes in scale. 
We thus propose adding scale augmentations during training, randomly resizing the search image by a factor of $s \in [0.5, 1.5]$.
While simple, we find this dramatically improves performance on \papername, improving STARK's AO by nearly $10\%$ and F-score by more than $8\%$ (\Cref{table:stark_tricks_comparison}).

\noindent \textbf{Context ratio.}
% In the tracking community, when extracting features from template image, they usually include some amount of background information. The most common practice is to use features within 2 times the size of the object. We experiment with different context ratio to see if this common is till the best in egocentric videos. However, because of the local window assumption, there is a relationship between the size of template image and size of search image: search\_img\_size / search\_region\_ratio = template\_img\_size / context\_ratio = object scale. When changing the context ratio, we need to carefully control other parameters for a fair comparison. The results are shown in \Cref{table:context_ratio}.
Past SOT works have found that including some background can be helpful for template image feature extraction, with twice the size of the object being common.
We experiment with different context ratios to see if this rule of thumb transfers to egocentric videos. 
Because of the local window assumption, the sizes of the template and search images are related: $\frac{\mathrm{Search~Image~Size (SIS)}}{\mathrm{Search~Region~Ratio (SRR)}} = \frac{\mathrm{Template~Image~Size}}{\mathrm{Context~Ratio (CR)}} = \mathrm{Object~Scale}$.
%  search\_img\_size / search\_region\_ratio = template\_img\_size / context\_ratio = object scale. 
The template image size is set to a fixed size $128\times 128$. When changing the context ratio, we carefully control the other parameters for a fair comparison. The results are shown in \Cref{table:context_ratio}.
Among all three parameters - \textbf{CR}, \textbf{SRR}, and \textbf{SIS}, the search region size (determined by \textbf{SRR} and \textbf{SIS}) has the highest impact on the F-score. This is expected because there are frequent re-detections, which require the tracker to search in a larger area for the object, rather than just within the commonly used local window. Varying the \textbf{CR} has mixed results so we adhere to the common practice of using a \textbf{CR} of 2. 
% The best result is achieved when using \textbf{SRR} 12.5, which covers most of the image and achieves a F-score of 43.74\%.

\vspace{-2mm}

\section{Future directions}
Based on our experiments in \Cref{table:sota_comparison} and \Cref{table:occ_comparison}, we found that re-detection to be a key challenge of long-term tracking, especially in egocentric video, where objects frequently go in and out of view, or are exposed to high motion blur. We see a few promising directions for future works:

a) Stronger features for associating objects should significantly improve re-detection; the impact of insufficiently discriminative feature embeddings can be clearly seen in the major gap in Tracking by Detection performance between the Oracle and InstEmb methods at the bottom of \Cref{table:sota_comparison}. Geometric keypoints, optical flow, or long-term trajectories \cite{wang2023tracking} can also lead to large improvements here.

b) Leveraging spatial signals: camera trajectories can be estimated as additional signals to the tracker. For example, if an object remains static during the window where it is out-of-view, knowledge of camera location can help re-localize the position of this object.

c) Global, multi-view object representations: Egocentric videos, with their diverse camera trajectories and tendency to capture the camera wearer’s interactions with objects, often offer significantly richer and more varied viewpoints of objects than traditional third-person tracking datasets. In the latter, object appearances tend to be more constant, so modern tracking methods have thus far been able to get away with using a single image template (optionally with an additional template from the latest frame). With a need for more robustness to the different viewpoints and occlusions offered by egocentric video, we believe that a challenging egocentric tracking dataset like EgoTracks represents an opportunity to develop trackers with more global, view-variant object representations learned in an online fashion. A simple version of this can be found in Section D of the supplementary material, where we augmented EgoSTARK by fusing multiple templates; we found that such a strategy indeed improved tracking results on EgoTracks.

\vspace{-1mm}
\section{Conclusion}
\vspace{-2mm}
We present \papername, the first large-scale dataset for long-term egocentric visual object tracking in diverse scenes. We conduct extensive experiments to understand the performance of state-of-the-art trackers on this new dataset, and find that they struggle considerably, possibly in part due to overfitting to some of the simpler characteristics of existing benchmarks.
% We thus propose several improvements to the STARK~\cite{yan2021learning} tracker, leading to a strong baseline that we call Ego-STARK, leading to vast improvements in performance on egocentric data. 
We thus propose several adaptations for the egocentric domain, leading to a strong baseline that we call Ego-STARK, which has vastly improved performance on EgoTracks. 
Lastly, we plan to organize a public benchmark challenge using a held-out test set with a test server as a testbed for new tracking algorithms. 
By publicly releasing this dataset and organizing the challenge, we hope to encourage advancements in the field of long-term tracking and draw more attention to the challenges of long-term and egocentric videos.

{\small
\bibliographystyle{plain}
\bibliography{main}
}

\section*{Checklist}
\begin{enumerate}

\item For all authors...
\begin{enumerate}
  \item Do the main claims made in the abstract and introduction accurately reflect the paper's contributions and scope?
    \answerYes{}
  \item Did you describe the limitations of your work?
    \answerYes{Please see Supplementary section Limitations.}
  \item Did you discuss any potential negative societal impacts of your work?
    \answerYes{Please see Supplementary section Potential negative societal impacts. The societal impact of this work is consistent with that of the Ego4D.}
  \item Have you read the ethics review guidelines and ensured that your paper conforms to them?
    \answerYes{}
\end{enumerate}

\item If you are including theoretical results...
\begin{enumerate}
  \item Did you state the full set of assumptions of all theoretical results?
    \answerNA{}
	\item Did you include complete proofs of all theoretical results?
    \answerNA{}
\end{enumerate}

\item If you ran experiments (e.g. for benchmarks)...
\begin{enumerate}
  \item Did you include the code, data, and instructions needed to reproduce the main experimental results (either in the supplemental material or as a URL)?
    \answerYes{Please see Supplementary Appendix A.}
  \item Did you specify all the training details (e.g., data splits, hyperparameters, how they were chosen)?
    \answerYes{}
	\item Did you report error bars (e.g., with respect to the random seed after running experiments multiple times)?
    \answerNo{Training/finetuning models is computationally costly to do multiple runs, similar to others in this field who do not report error bars. For all other baselines, we use their official checkpoints for inference only, so their results are deterministic.}
	\item Did you include the total amount of compute and the type of resources used (e.g., type of GPUs, internal cluster, or cloud provider)?
    \answerYes{} See Section~\ref{sec:ego_design}.
\end{enumerate}

\item If you are using existing assets (e.g., code, data, models) or curating/releasing new assets...
\begin{enumerate}
  \item If your work uses existing assets, did you cite the creators?
    \answerYes{} Our annotations are on top of Ego4D~\cite{grauman2022ego4d}, which we discuss at length.
  \item Did you mention the license of the assets?
    \answerYes{}
  \item Did you include any new assets either in the supplemental material or as a URL?
    \answerYes{Please see Supplementary Appendix A.}
  \item Did you discuss whether and how consent was obtained from people whose data you're using/curating?
    \answerYes{} See Section~\ref{sec:ego4d_vq}; the videos in Ego4D~\cite{grauman2022ego4d} were collected by consenting participants.
  \item Did you discuss whether the data you are using/curating contains personally identifiable information or offensive content?
    \answerYes{} See Section~\ref{sec:ego4d_vq}; by annotating Ego4D~\cite{grauman2022ego4d}, we inherit its de-identification and screening for offensive content.
\end{enumerate}

\item If you used crowdsourcing or conducted research with human subjects...
\begin{enumerate}
  \item Did you include the full text of instructions given to participants and screenshots, if applicable?
    \answerYes{} We include detailed instructions in the supplementary. For proprietary reasons, we are not able to provide screenshots of the annotation tools.
  \item Did you describe any potential participant risks, with links to Institutional Review Board (IRB) approvals, if applicable?
    \answerNA{} There are no human studies done in this paper. However, we build on Ego4D~\cite{grauman2022ego4d}, which was collected with IRB approval.
  \item Did you include the estimated hourly wage paid to participants and the total amount spent on participant compensation? 
    \answerNA{} The participants were employed by a third-party vendor and are compensated based on the agreement with their employer.
\end{enumerate}

\end{enumerate}

%%%%%%%%%%%%%%%%%%%%%%%%%%%%%%%%%%%%%%%%%%%%%%%%%%%%%%%%%%%%

\end{document}